\documentclass[letterpaper,10pt,conference]{ieeeconf}

\IEEEoverridecommandlockouts
\usepackage{amsmath}
\usepackage{amssymb}
\usepackage{booktabs}
\usepackage{graphicx}
\usepackage{multirow}
\usepackage{subcaption}
\usepackage{tikz}
\usepackage{tikz-3dplot}
\usepackage{url}
\usepackage{xcolor}

\usetikzlibrary{
    arrows.meta,
    calc,
    positioning
}

\begin{document}

\title{\LARGE\bfseries
Contact-Aided Factor-Graph Localization\\
for Underwater Sampling
}

\author{
Michele Grimaldi$^{1}$,
Yosaku Maeda$^{2}$,
Hitoshi Kakami$^{2}$,\\
Ignacio Carlucho$^{1}$,
Yvan R. Petillot$^{1}$,
and Tomoya Inoue$^{2}$%
\thanks{$^{1}$School of Engineering \& Physical Sciences,
Heriot-Watt University, Edinburgh, UK.
{\tt\small m.grimaldi@hw.ac.uk}}%
\thanks{$^{2}$Engineering Department,
Japan Agency for Marine-Earth Science and Technology,
Yokosuka, Japan.}%
}

\maketitle
\thispagestyle{empty}
\pagestyle{empty}

\begin{abstract}
Accurate state estimation for autonomous underwater vehicles performing close-range seafloor sampling remains challenging. 
%
In low-altitude operation, down-looking cameras over featureless planar seabeds produce scale ambiguity, lateral degeneracy, and inconsistent feature tracking. Meanwhile, inertial–Doppler Velocity Log (DVL) fusion alone provides no mechanism for structural drift correction. 
We propose a Contact-Aided Factor-Graph Localization framework that treats physical interaction as an informative geometric constraint within a smoothing-based localization formulation. The method tightly fuses suction-based manipulator contact events with adaptive visual odometry, learned object detections and on-board sensors. Visual odometry relative-pose factors and landmark bearing-range factors are uncertainty-scaled according to inlier statistics to prevent visually weak frames from destabilizing the estimator, while contact events are modeled as high-confidence factors that induce implicit loop closures without appearance-based place recognition.
Furthermore, the system can fully initialize online during motion.
Experimental evaluation in tanks, harbor, and simulation environments demonstrates that contact-induced constraints significantly reduce trajectory drift and improve object revisit accuracy compared to filtering-based navigation and contact-free graph formulations. These results highlight the role of embodied physical interaction as a localization primitive in perception-degraded underwater environments.
\end{abstract}

\begin{figure*}
    \centering
    \includegraphics[width=0.8\linewidth]{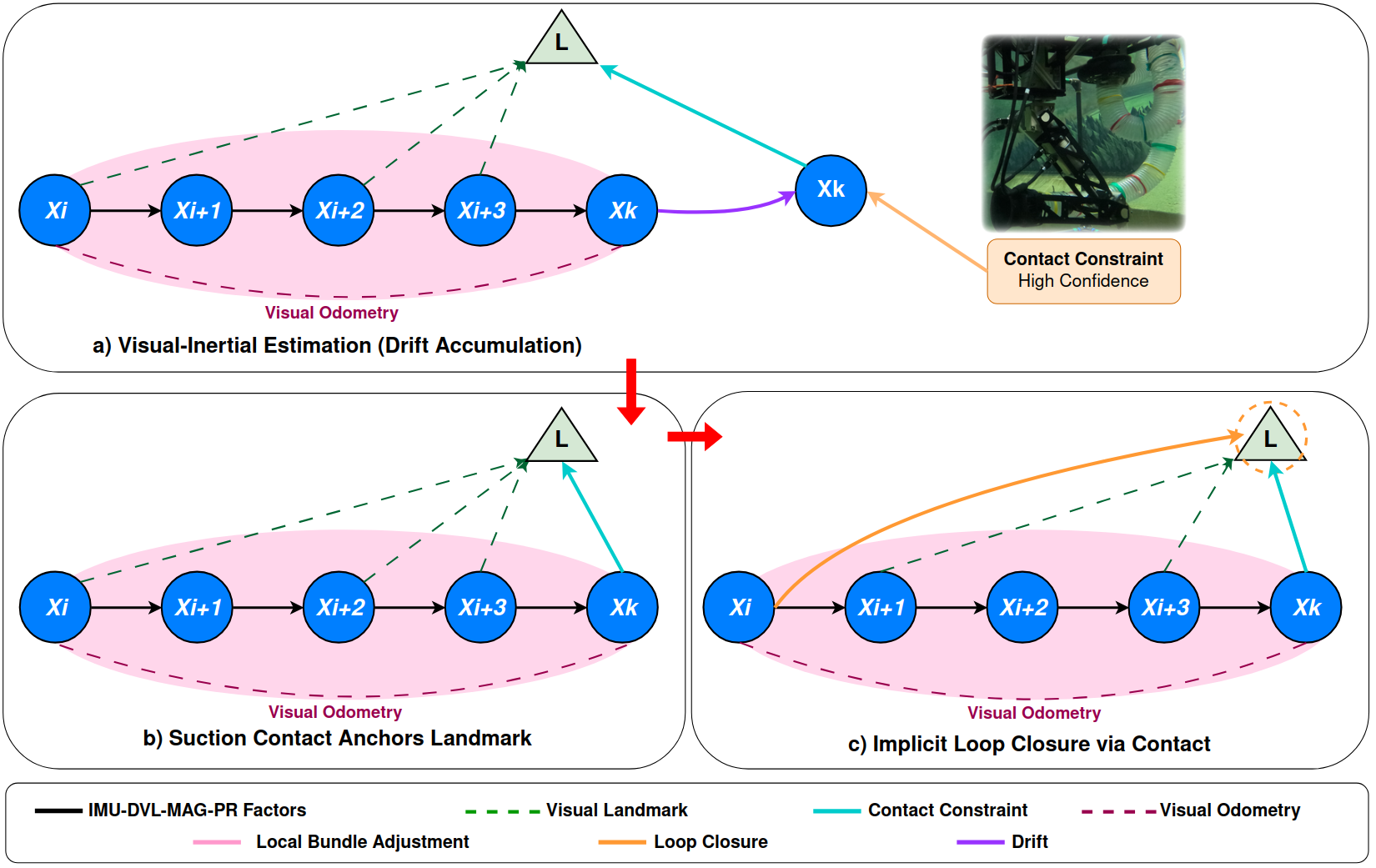}
\caption{
\textbf{Suction-based implicit loop closure.}
(a) Visual--inertial odometry accumulates drift over time despite continuous visual and inertial constraints.
(b) When the manipulator establishes suction contact, the contact point is modeled as a fixed landmark, anchoring the current pose with a high-confidence constraint.
(c) This contact induces an implicit loop closure in the factor graph, propagating corrections through the trajectory and reducing global drift.
}
    \label{fig:teaser}
    \vspace{-1em}
\end{figure*}


\section{Introduction}

Autonomous underwater vehicles (AUVs) performing close-range seafloor sampling operate in severely perception-degraded conditions. GPS is unavailable, and navigation relies on inertial–DVL dead reckoning, which accumulates drift. Although visual SLAM has shown strong performance in terrestrial robotics, underwater imagery suffers from scattering, turbidity, low texture, and photometric distortion \cite{ferrera, Grimaldi_2023}. 
These limitations are amplified in low-altitude, down-looking configurations typical of benthic sampling. The scene is predominantly planar and texture-sparse, rendering monocular visual odometry geometrically ill-conditioned: translation parallel to the seabed is weakly observable, metric scale is ambiguous without reliable altitude cues, and feature tracking is unstable. 
While visual landmarks can provide constraints, benthic organisms are small, sparsely distributed, and only briefly visible. Monocular range estimates are noisy, and re-identification is unreliable, yielding intermittent constraints that do not ensure persistent global consistency during manipulation. Accurate localization, however, is essential for object approach, suction-based grasping, and repeated sampling. The core challenge is therefore how to obtain structurally informative constraints in environments where conventional exteroceptive sensing is weak.
We propose to treat physical interaction as a localization primitive. When the AUV establishes suction contact with a benthic object, the end-effector creates a metrically grounded, viewpoint-invariant geometric constraint. We model these contact events as factors within a smoothing-based SLAM framework, allowing contact to induce implicit loop closures without appearance-based place recognition, as illustrated in Fig.~\ref{fig:teaser}. In perception-degraded underwater settings, physical interaction becomes a mechanism for drift mitigation and global consistency. We introduce a Contact-Aided Factor-Graph Localization framework that tightly fuses IMU, DVL, pressure, and magnetometer measurements with adaptive down-looking visual odometry, learned object detections, and manipulator contact constraints. The system addresses three challenges:

\begin{enumerate}
    \item \textbf{Planar visual degeneracy:} Adaptive uncertainty scaling mitigates estimator inconsistency under weak monocular geometry.
    \item \textbf{Drift without natural loop closures:} Suction contact is formulated as a high-confidence geometric factor that induces implicit loop closure.
    \item \textbf{Intermittent visual structure:} A hybrid sliding-window strategy alternates between landmark-augmented and pose-only refinement to preserve real-time robustness.
\end{enumerate}

The system initializes fully online and runs incrementally using an iSAM2-based backend \cite{5979641}. We validate the approach in physics-based simulation, tanks experiments, and harbor deployments. Results show that contact-induced constraints significantly reduce drift and improve object revisit accuracy compared to filtering-based navigation and contact-free graph formulations. Our main contributions are: 
\begin{itemize}
    \item  We formalize suction-based contact as a structurally informative factor that resolves observability limitations in low-texture underwater environments. 
    \item We develop a tightly coupled contact-aided smoothing framework for low-altitude AUV sampling missions. 
    \item We introduce degeneracy-aware visual uncertainty modeling and a hybrid real-time optimization strategy for intermittent visual structure.
\end{itemize}

Together, these results demonstrate that deliberate physical interaction can substantially enhance localization robustness in perception-limited underwater domains.

\section{Related Work}

Underwater state estimation is inherently more challenging than terrestrial SLAM due to the absence of GPS, limited visibility, and degraded proprioceptive sensing. Classical AUV navigation systems fuse inertial, DVL, and depth measurements using filtering or graph-based formulations to reduce dead-reckoning drift~\cite{Kinsey2006ASO}. However, without reliable loop closures or strong exteroceptive cues, drift remains unavoidable. Visual SLAM has been widely explored in underwater robotics due to its low cost and rich environmental information. Empirical studies and surveys report substantial performance degradation underwater, caused by scattering, turbidity, low texture, and illumination variability~\cite{Grimaldi_2023, rs15102496}. These effects result in unstable feature tracking, ambiguous relative pose estimates, and unreliable loop closure detection. The problem is exacerbated in low-altitude, down-looking configurations, where planar and repetitive seafloor imagery leads to geometric degeneracy and weak observability of lateral motion. Approaches to mitigate these issues include image enhancement, adaptive residual weighting, and tighter visual–inertial–depth coupling~\cite{11007055,DING2024115245}. While these methods improve robustness, they remain fundamentally exteroceptive and cannot guarantee persistent global consistency when visual structure is weak or intermittent. To further enhance robustness, several systems integrate additional sensing modalities. Tightly coupled acoustic–visual–inertial frameworks combine DVL, stereo cameras, and IMUs to operate under degraded visibility~\cite{xu2025aquaslamtightlycoupledunderwateracousticvisualinertial}, and systems such as SVIn2 incorporate sonar, visual, inertial, and depth measurements with loop closure and relocalization capabilities~\cite{8967703}. Although multi-modal fusion improves accuracy and consistency, these approaches still depend on environmental perceptual structure and are therefore susceptible to observability limitations in texture-sparse benthic environments~\cite{11128101}.
Contact-based SLAM has been explored in terrestrial robotic manipulation, where tactile or force feedback augments state estimation under visual occlusion~\cite{wang2025contactslamactivetactile}. These works demonstrate that physical interaction can provide informative geometric constraints. However, contact has not been leveraged as a localization primitive in underwater SLAM. In contrast, we introduce a Contact-Aided Factor-Graph Localization framework that integrates suction-based contact constraints with adaptive visual and inertial measurements within a unified smoothing formulation. By modeling physical interaction as a metrically grounded factor that induces implicit loop closure, the proposed method directly addresses structural observability limitations in low-texture, down-looking underwater environments.

\section{Methodology}

Our framework formulates underwater state estimation as a contact-aided smoothing problem on a factor graph, implemented using GTSAM~\cite{gtsam, factor_graphs_for_robot_perception}. Vehicle poses, velocities, IMU biases, and environmental landmarks are jointly estimated by fusing proprioceptive navigation measurements, down-looking visual observations, learned object detections, and manipulator contact constraints within a unified probabilistic framework. The estimator initializes fully online, without requiring stationarity or a dedicated bootstrapping phase. As soon as IMU, magnetometer, pressure, and DVL data are available, the initial pose is constructed from roll–pitch–yaw and pressure-derived depth, velocity is initialized from DVL measurements, and Gaussian priors are applied to pose, velocity, and bias states. 

\subsection{Inertial Sensors}

The AUV integrates IMU, DVL, pressure sensor, and magnetometer measurements within a factor-graph framework to estimate 6-DOF pose and velocity. The IMU provides body-frame acceleration $\mathbf{a}_b$ and angular velocity $\boldsymbol{\omega}_b$. Measurements are rotated to the vehicle frame via $\mathbf{R}_{\text{imu}}$:

\begin{equation}
\mathbf{a}_v = \mathbf{R}_{\text{imu}} \mathbf{a}_b, \quad
\boldsymbol{\omega}_v = \mathbf{R}_{\text{imu}} \boldsymbol{\omega}_b.
\end{equation}

Using standard preintegration over $\Delta t$, relative motion increments are computed as:

\begin{equation}
\begin{aligned}
\Delta \mathbf{R}_{k+1} &= \Delta \mathbf{R}_k \exp((\boldsymbol{\omega}_v - \mathbf{b}_\omega)\Delta t), \\
\Delta \mathbf{v}_{k+1} &= \Delta \mathbf{v}_k + (\mathbf{a}_v - \mathbf{b}_a)\Delta t, \\
\Delta \mathbf{p}_{k+1} &= \Delta \mathbf{p}_k + \Delta \mathbf{v}_k \Delta t + \tfrac{1}{2}(\mathbf{a}_v - \mathbf{b}_a)\Delta t^2.
\end{aligned}
\end{equation}

Biases $\mathbf{b}_a$ and $\mathbf{b}_\omega$ are jointly estimated. The resulting preintegrated IMU factor links consecutive states $(\mathbf{x}_k, \mathbf{v}_k, \mathbf{b}_k)$ and $(\mathbf{x}_{k+1}, \mathbf{v}_{k+1}, \mathbf{b}_{k+1})$:

\begin{equation}
f_{\text{IMU}} \sim \mathcal{N}(\Delta \mathbf{p}_{k+1}, \Sigma_{\text{IMU}}).
\end{equation}

The DVL provides body-frame velocity $\mathbf{v}_{\text{dvl}}$, corrected for rotational motion using lever-arm offset $\mathbf{r}_{\text{IMU→DVL}}$:
\begin{equation}
\mathbf{v}_c = \mathbf{v}_{\text{dvl}} - \boldsymbol{\omega}_v \times \mathbf{r}_{\text{IMU→DVL}},
\quad
\mathbf{v}_w = \mathbf{R}_{\text{imu}} \mathbf{v}_c.
\end{equation}

This is incorporated as a velocity factor:
\begin{equation}
f_{\text{DVL}}(\mathbf{v}_k) \sim \mathcal{N}(\mathbf{v}_w, \Sigma_{\text{DVL}}).
\end{equation}

Pressure measurements are converted to depth using a 
temperature-compensated density model,
\[
z = \frac{P - P_0}{(\rho - 0.2\,T)\, g},
\]
where $P$ is the absolute pressure, $P_0$ atmospheric pressure,
$T$ the water temperature, $\rho$ the nominal water density,
and $g$ gravity. The depth is fused as a prior on the vertical position:
\begin{equation}
f_{\text{Pressure}}(\mathbf{x}_k)
\sim
\mathcal{N}(z, \sigma_z^2).
\end{equation}

The magnetometer provides a yaw measurement $\psi_m$ (heading about the
gravity axis), incorporated as:
\begin{equation}
f_{\text{Mag}}(\mathbf{x}_k)
\sim
\mathcal{N}(\psi_m, \sigma_\psi^2),
\end{equation}
where $\sigma_\psi^2$ denotes the yaw variance.
Together, these factors provide tightly coupled inertial navigation, constraining inter-node motion, horizontal velocity, vertical position, and heading. This fused inertial backbone ensures stable pose estimation when visual structure is weak or intermittent.

\subsection{Down-Looking Visual Odometry}

A down-looking monocular camera provides relative motion constraints from the seafloor. In planar, low-texture, low-altitude operation, monocular VO is inherently noisy and scale-ambiguous. We therefore employ a tightly coupled pipeline with multi-sensor scale recovery, lever-arm compensation, and adaptive uncertainty modeling. Images are contrast-enhanced (CLAHE) \cite{109340}. ORB features \cite{Campos_2021} are extracted with BRISK fallback \cite{6126542}, and matched using Hamming distance with a ratio test. The essential matrix $\mathbf{E}$ is estimated via RANSAC and decomposed to obtain relative rotation $\mathbf{R}_c$ and translation direction $\hat{\mathbf{t}}_c$:
\[
(\mathbf{R}_c, \hat{\mathbf{t}}_c) = \text{recoverPose}(\mathbf{E}).
\]

Metric scale $s$ is recovered from auxiliary sensors: DVL or sonar altitude when available, otherwise pressure-derived depth, with IMU-based fallback. The scaled motion is $\mathbf{t}_c = s \hat{\mathbf{t}}_c$, forming the relative camera pose $\mathbf{T}_c^{k-1,k} = (\mathbf{R}_c, \mathbf{t}_c)$. Using fixed extrinsics $\mathbf{T}_{cb}$, this is transformed to the body frame:
\begin{equation}
\mathbf{T}_b^{k-1,k} = \mathbf{T}_{cb}^{-1} \mathbf{T}_c^{k-1,k} \mathbf{T}_{cb}.
\end{equation}

The resulting motion is added as a BetweenFactor:
\begin{equation}
f_{\text{VO}}(\mathbf{x}_{k-1}, \mathbf{x}_k) \sim \mathcal{N}(\mathbf{T}_b^{k-1,k}, \Sigma_{\text{VO}}).
\end{equation}

A corresponding velocity constraint is formed from the scaled translation and corrected for the camera–IMU lever arm $\mathbf{r}_{\text{cam}\rightarrow\text{imu}}$:
\begin{equation}
\mathbf{v}_{\text{corr}} = \frac{\mathbf{t}_b}{\Delta t} - \boldsymbol{\omega} \times \mathbf{r}_{\text{cam}\rightarrow\text{imu}},
\end{equation}
and incorporated as
\begin{equation}
f_{\text{VO-vel}}(\mathbf{v}_{k-1}, \mathbf{v}_k) \sim \mathcal{N}(\mathbf{v}_{\text{corr}}, \Sigma_{\text{VO-vel}}).
\end{equation}

To mitigate visually unreliable updates, covariance is adaptively scaled using the inlier ratio: 
\begin{equation}
\eta = \frac{N_{\text{inliers}}}{N_{\text{matches}}},
\end{equation}
smoothed via an exponential moving average and clamped within bounded limits:
\begin{equation}
\Sigma_{\text{VO}} = \alpha_k \,\text{diag}(\sigma_t^2, \sigma_r^2).
\end{equation}

This formulation preserves informative visual constraints when structure is available while downweighting degenerate observations, enabling robust integration of down-looking VO within the factor-graph framework.

\subsection{Visual Landmark Detections}

Object detections are produced by a YOLOv8-based neural network running on the down-looking camera stream. The detector is trained on a public starfish dataset\footnote{\url{https://www.kaggle.com/datasets/yusufsyam/starfish}} and provides, for each detection, a class label $c$, pixel center $(u,v)$, a confidence score, and a bounding box. A slant range $r$ is estimated from the bounding-box height $h$ using a pinhole model and an assumed object size $D$, such that $
r = \frac{f_y D}{h}$,
and is bounded by the DVL altitude when available. The pixel center is back-projected into a normalized camera-frame bearing,
\[
\mathbf{b}_c = 
\frac{1}{\|\mathbf{d}_c\|}\mathbf{d}_c, 
\quad 
\mathbf{d}_c = 
\begin{bmatrix}
(u-c_x)/f_x & (v-c_y)/f_y & 1
\end{bmatrix}^\top.
\]

Using fixed camera–body extrinsics $\mathbf{T}_{cb}$ and the current pose $\mathbf{T}_w^b(k)$, detections are transformed into the world frame and associated with persistent landmarks. Each detection contributes a bearing–range factor,
\[
f_{\text{BR}}(\mathbf{x}_k,\mathbf{l}_j)
\sim
\mathcal{N}\!\bigl((\mathbf{b}_b, r), \sigma^2\mathbf{I}_3\bigr),
\]
where the measurement uncertainty $\sigma$ is inflated for low-confidence or small bounding-box detections. This formulation enables incremental landmark creation while maintaining geometric consistency between neural detections and vehicle pose. Landmarks are also expressed in the end-effector frame, allowing direct anchoring of visual detections to contact events for contact-consistent mapping and implicit loop closure.

\subsection{Hybrid Local Bundle Adjustment}

To improve short-term consistency while preserving real-time performance, we employ a sliding-window Local Bundle Adjustment (LBA) over the most recent poses 
$\{\mathbf{x}_{k-N+1}, \ldots, \mathbf{x}_k\}$.
At each invocation, the factor graph is queried to determine whether visual landmarks are connected to any pose within the window. If so, a landmark-augmented LBA is performed; otherwise, the system falls back to pose-only refinement.

\paragraph{Landmark-Augmented Optimization.}
All factors connected to window poses are extracted to form a local subgraph with active variables $\mathcal{X}_{\text{local}}$ (poses and associated landmarks) and factors $\mathcal{F}_{\text{local}}$. The optimization solves
\begin{equation}
\mathcal{X}_{\text{local}}^\star =
\arg\min_{\mathcal{X}_{\text{local}}}
\sum_{f_i \in \mathcal{F}_{\text{local}}}
\| f_i(\mathcal{X}_{\text{local}}) \|_{\Sigma_i}^2.
\end{equation}

To prevent gauge freedom, poses outside the window but connected to the subgraph are anchored with strong priors at their current estimates. The resulting nonlinear least-squares problem is solved using bounded Levenberg–Marquardt iterations, and optimized values are reintegrated into the global estimate.

\paragraph{Pose-Only Fallback.}
If no landmarks are present, only pose-related factors $\mathcal{F}_{\text{pose}}$ and variables $\mathcal{X}_{\text{pose}}$ are retained:
\begin{equation}
\mathcal{X}_{\text{pose}}^\star =
\arg\min_{\mathcal{X}_{\text{pose}}}
\sum_{f_i \in \mathcal{F}_{\text{pose}}}
\| f_i(\mathcal{X}_{\text{pose}}) \|_{\Sigma_i}^2.
\end{equation}

This adaptive switching preserves robustness in visually sparse conditions while exploiting landmark constraints when available. The sliding-window refinement reduces short-term drift and improves local smoothness without compromising incremental iSAM2 performance.

\subsection{Manipulator-Based Contact Sensing}

The vehicle-mounted manipulator is treated as an additional proprioceptive sensor that provides contact-based geometric constraints through a suction end-effector. 
It is actuated using an acceleration-level inverse-kinematics controller \cite{10753793_real,grimaldi2026reliablesubseaobjectrecovery}, ensuring smooth, precise end-effector positioning during contact events and suction-based interactions.
Forward kinematics from joint encoders yields the end-effector pose in the world frame,
\[
\mathbf{T}_w^{ee}(k) = \mathbf{T}_w^b(k)\,\mathbf{T}_b^{ee}(\boldsymbol{\theta}_k),
\]
where $\boldsymbol{\theta}_k$ are the arm joint angles. When the suction cup engages, a rising-edge event is detected and the current position of the end-effector $\mathbf{p}_{ee}$ is calculated. This contact event is treated as a point measurement in the environment. If an existing landmark $\mathbf{l}_j$ lies within a proximity threshold,
($
\|\mathbf{l}_j - \mathbf{p}_{ee}\| < \epsilon$),
a bearing–range factor is added between the current pose $\mathbf{x}_k$ and the landmark,



\begin{equation}
\begin{aligned}
f_{\text{contact}} &(\mathbf{x}_k,\mathbf{l}_j)
\sim
\mathcal{N}\!\bigl((\hat{\mathbf{b}}, d), \Sigma_c\bigr), \\
& \hat{\mathbf{b}} = \frac{\mathbf{l}_j - \mathbf{p}_{ee}}{\|\mathbf{l}_j - \mathbf{p}_{ee}\|},\\ 
& d = \|\mathbf{l}_j - \mathbf{p}_{ee}\|.
\end{aligned}
\end{equation}
If no nearby landmark exists, a new landmark is initialized at the contact point, 
$\mathbf{l}_{new} = \mathbf{p}_{ee}$,
and a zero-range contact factor is added. In both cases, a tight isotropic noise model $\Sigma_c$ enforces strong geometric consistency. This mechanism yields an implicit loop closure when the end-effector contacts a previously mapped object, without requiring explicit place recognition. The manipulator thus acts as an active tactile sensor, providing sparse but high-confidence constraints that stabilize the factor graph during prolonged low-visibility operation.

\section{Experimental Setup}

\begin{figure}[t]
     \centering
     \includegraphics[width=\linewidth]{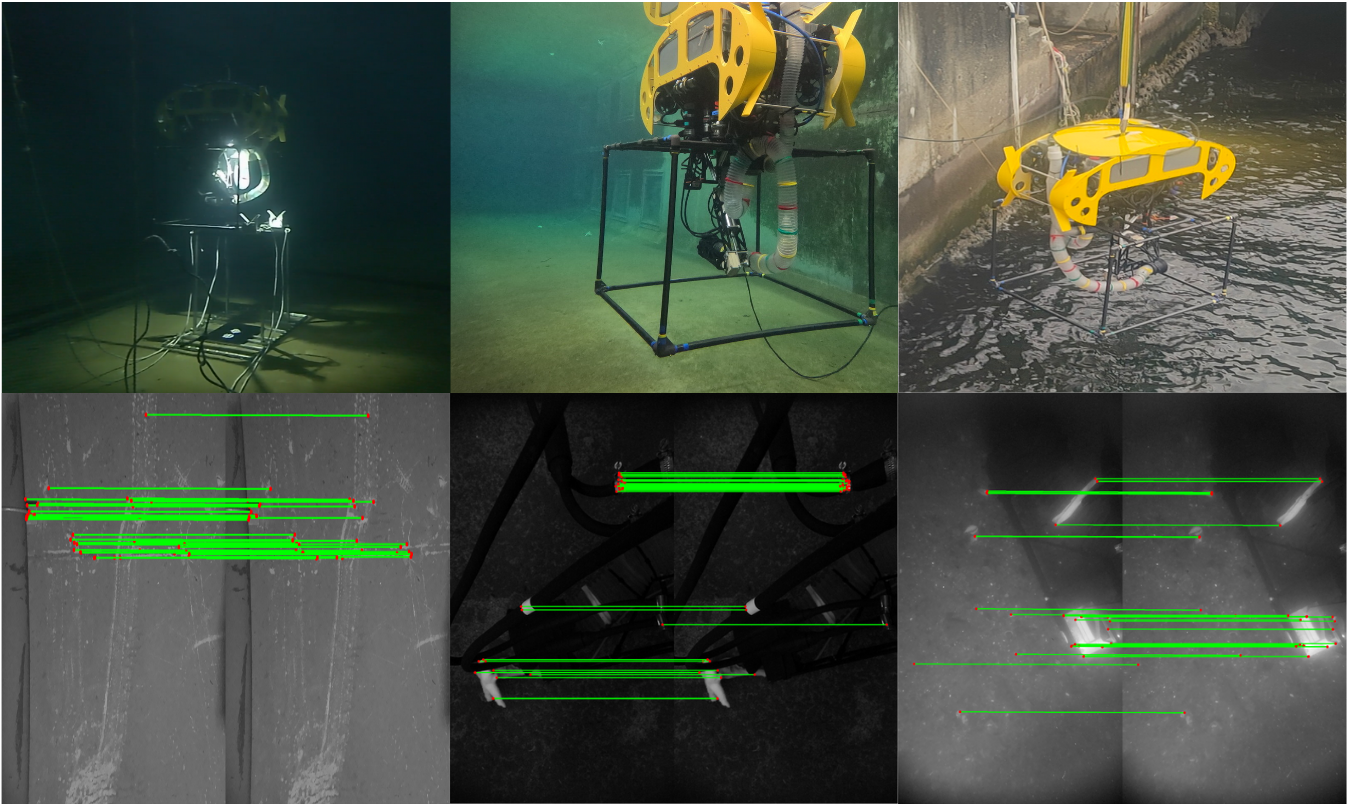}
\caption{Experimental environments. Top: first tank (low light), second tank (natural sunlight), and harbor deployment site. Bottom: representative best-match image pairs illustrating the geometric degeneracy and impracticality of classical Structure-from-Motion and monocular visual SLAM in these conditions.}
\label{fig:exp_setup}
\vspace{-1.5em}
\end{figure}


\begin{figure*}[t]
    \centering
    \includegraphics[width=\linewidth]{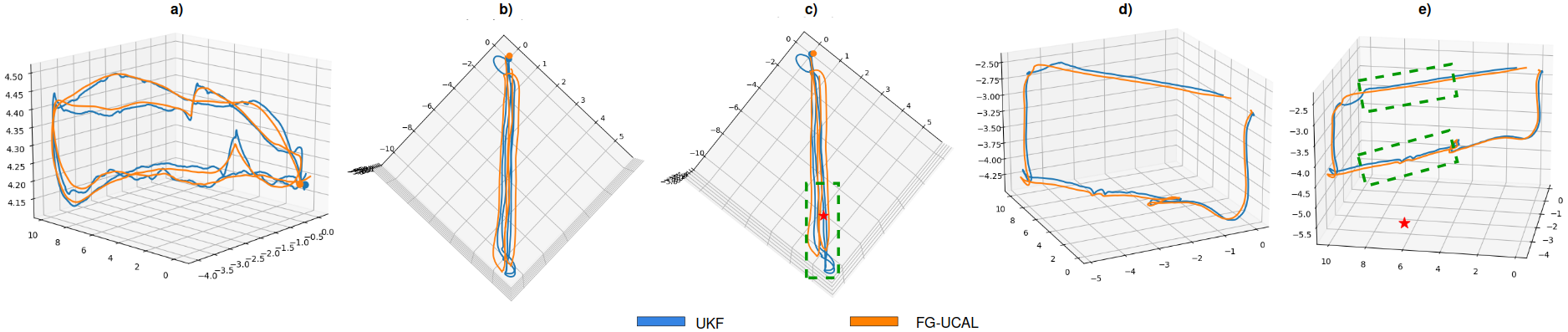}
    \caption{\textbf{First tank experiments}: (a) T1: straight trajectory; (b) T2: two 180$^\circ$ rotations; (c) T3: rotations with starfish landmark; (d) T4: rotations with depth variation; (e) T5: depth variation and starfish landmark. Red markers denote the starfish target, and dashed green lines indicate segments with improved estimation accuracy.}
    \label{fig:trajs_tank1}
    \vspace{-0.45cm}
\end{figure*}

The proposed framework is evaluated in simulation and real-world deployments. Simulation experiments are conducted in Stonefish \cite{grimaldi2025stonefishsupportingmachinelearning}, which provides perfect ground-truth odometry for quantitative validation. Real-world trials are performed in two controlled water tanks and one harbor site, shown in Fig.~\ref{fig:exp_setup}. The tanks measure 100\,m $\times$ 7.8\,m $\times$ 4.35\,m and 40\,m $\times$ 4\,m $\times$ 2\,m, while the harbor trials cover approximately 10\,m $\times$ 10\,m $\times$ 5\,m under natural disturbances including turbidity and illumination variability. Trajectories are designed to stress observability in planar, low-texture conditions. In the larger tank, we execute (i) straight motion without rotation, (ii) motion with two 180$^\circ$ yaw turns, and (iii) combined yaw rotations with depth variation. Additional trials include repeated approaches to a benthic target (starfish) to evaluate revisit consistency under manipulation-like motion. Simulation experiments replicate representative motion patterns, including continuous rotations and interaction sequences, enabling direct comparison against ground truth. The vehicle is equipped with a JIMS-80 MEMS IMU, Wayfinder DVL, Paroscientific 8BT11000-I pressure and temperature sensor, magnetometer, and a downward-looking monocular camera. IMU, DVL, pressure, and magnetometer measurements are sampled at 10\,Hz, while the camera operates at 2\,Hz. IMU noise parameters are identified via Allan variance analysis using over 9\,h of stationary data~\cite{5570702}.

\begin{table*}[t]
\centering
\footnotesize
\caption{Trajectory accuracy metrics from the first tank experiments. 
The upper section reports full-system performance across different motion profiles. 
The lower section presents the ablation study conducted on Trajectory~T1 (No Rotation).}
\label{tab:trajectory_and_ablation}
\begin{tabular}{lcccc}
\hline
Configuration / Trajectory & ATE (m) & RPE (m) & Yaw RMSE (rad) & Yaw RPE (rad) \\
\hline
\multicolumn{5}{c}{\textit{Full System -- Different Motion Profiles}} \\
\hline
T1: No Rotation
& 0.0990 & 0.0812 & 0.0524 & 0.0113 \\

T2: $2\times180^\circ$ Rotations
& 0.2917 & 0.0747 & 0.1232 & 0.0627 \\

T3: $2\times180^\circ$ Rotations + Starfish Landmark
& 0.3514 & 0.0812 & 0.0867 & 0.0905 \\

T4: $2\times180^\circ$ Rotations + Depth Change
& 0.1954 & 0.0489 & 0.0639 & 0.0134 \\

T5: $2\times180^\circ$ Rotations + Depth Change + Starfish Landmark
& 0.1651 & 0.0673 & 0.0412 & 0.0097 \\

\hline
\multicolumn{5}{c}{\textit{Ablation Study -- Trajectory T1 (No Rotation)}} \\
\hline
FG-UCAL (LBA, 10-Pose Window)
& 0.1383 & 0.0168 & 0.0790 & 0.0035 \\

FG-UCAL (LBA, 25-Pose Window)
& 0.1629 & 0.0168 & 0.0920 & 0.0041 \\

FG-UCAL (IMU Preintegration: 2 Samples)
& 0.1740 & 0.0167 & 0.0925 & 0.0037 \\

FG-UCAL (No Local Bundle Adjustment)
& 0.2270 & 0.0169 & 0.0929 & 0.0041 \\

FG-UCAL (No Camera Updates)
& 0.2798 & 0.0281 & 0.1391 & 0.0077 \\ 

DVL--IMU--Pressure 
& 0.6780 & 0.0182 & 2.2396 & 0.0033 \\

DVL--IMU--Pressure (with LBA)
& 0.6763 & 0.0181 & 2.3187 & 0.0033 \\

\hline
\end{tabular}
\end{table*}

\subsection{Visual Degeneracy Characterization}

All environments exhibit predominantly planar, low-texture seabed structure, creating severe geometric degeneracy for monocular reconstruction. Quantitative feature statistics confirm this limitation. In the harbor sequences, an average of $418$ keypoints per frame are detected (minimum $90$), with only $42\%$ retained between frames. The first tank yields higher detections (mean $781.8$), yet only $47\%$ survive temporally, indicating weak feature persistence despite increased raw detections. The second tank further highlights the degeneracy. Across $4078$ analyzed frames, an average of $481.2$ keypoints per frame are detected (minimum $53$, maximum $1000$), with $288.0$ matches per frame on average (minimum $28$, maximum $688$). The mean tracking ratio (matches/keypoints) is $0.59$ (minimum $0.25$, maximum $0.79$). Although this ratio appears moderate, the optical flow median magnitude remains very small (mean $1.24\,\mathrm{px}$), indicating insufficient parallax for stable depth triangulation. The gradient magnitude is similarly low (mean $15.60$), reflecting limited photometric structure. Across all sequences, gradient magnitudes remain in the range $9.38$--$15.60$, and optical flow is either insufficient for meaningful parallax ($\approx 1.24\,\mathrm{px}$) or inconsistent (up to $31.22\,\mathrm{px}$), leading to unstable depth estimation. Under these conditions, COLMAP~\cite{schoenberger2016mvs,schoenberger2016sfm} fails to initialize reliable sparse reconstructions. ORB-SLAM3, VINS-MONO, and ROVIO~\cite{rovio} repeatedly lose tracking or diverge. Figure~\ref{fig:exp_setup} illustrates representative best-match pairs, highlighting the absence of stable multi-view geometry. These results empirically demonstrate that classical monocular SfM and visual SLAM pipelines are fundamentally ill-conditioned in this regime.


\begin{figure*}[t]
    \centering
    \includegraphics[width=\linewidth]{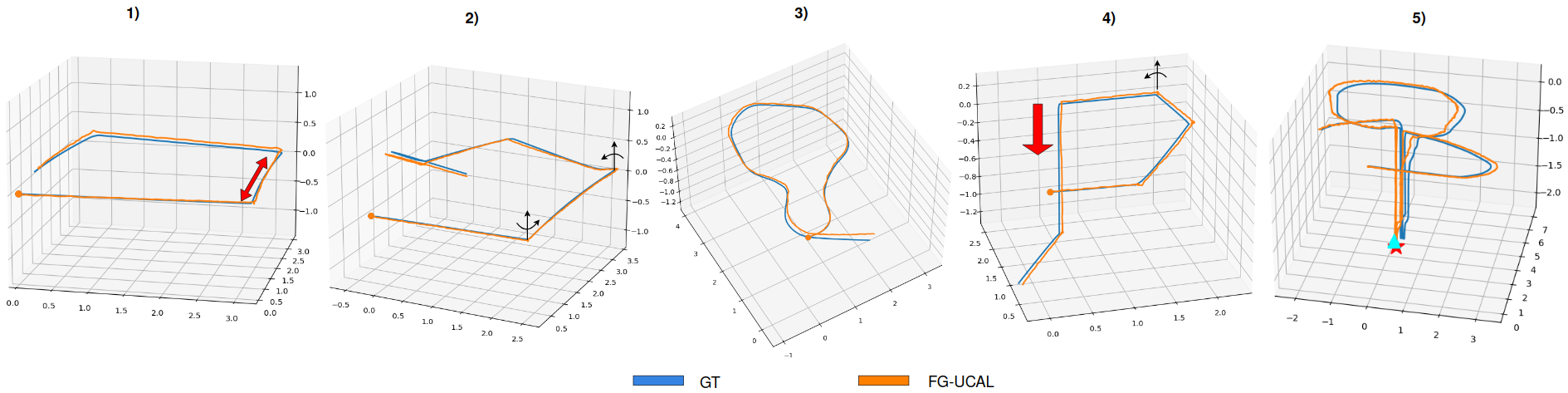}
    \caption{Stonefish simulation experiments. Estimated trajectories using the proposed FG-UCAL framework are compared against ground-truth (perfect odometry) for representative motion patterns: S1: forward–lateral motion, S2: multiple rotations, S3: continuous rotations, S4: depth with yaw variation, and a S5: semi-8 trajectory with two starfish dives.}
    \label{fig:stonefish_trajs}
\end{figure*}

\begin{table*}[t]
\centering
\caption{Trajectory error metrics in Stonefish simulation relative to ground-truth odometry. ATE and RPE are reported in meters; angular errors in radians.}
\setlength{\tabcolsep}{4pt}
\begin{tabular}{lcccc}
\hline
Trajectory & ATE (m) & RPE (m) & Yaw RMSE (rad) & Yaw RPE (rad) \\
\hline

S1: Forward and lateral movement
& 0.0903 & 0.0039 
& 0.0538 & 0.0039 \\

S2: Multiple rotations 
& 0.1635 & 0.0060 
& 0.0315 & 0.0025 \\

S3: Continuous rotations
& 0.998 & 0.0032 
& 0.0886 & 0.0415 \\

S4: Depth + yaw change 
& 0.0847 & 0.0052 
& 0.0981 & 0.0080 \\

S5: Semi 8 shape + 2 starfish dives
& 0.2743 & 0.0178 
& 0.1290 & 0.0561 \\
\quad $\hookrightarrow$ Landmark and Pump – Starfish revisit distance (m)
& \multicolumn{4}{c}{0.06 m} \\

\hline
\end{tabular}
\label{tab:trajectory_summary}
\vspace{-1em}
\end{table*}

\section{Results}

We evaluate the proposed framework across two controlled tank environments and a real harbor deployment. The experiments are designed to assess (i) estimator behavior under planar degeneracy, (ii) the contribution of landmarks and contact events, and (iii) task-level revisit consistency during repeated benthic interactions.

\subsection{First Tank: Controlled Observability Analysis}

The first tank experiments isolate observability effects under structured motion, including straight translation, repeated $180^\circ$ yaw rotations, rotations with a starfish landmark, depth variation, and their combination. 
Fig.~\ref{fig:trajs_tank1} shows the five trajectories used, where red markers denote the starfish target, and dashed green segments indicate trajectory portions where additional geometric excitation improves estimation. Quantitative results are summarized in Table~\ref{tab:trajectory_and_ablation}. Under straight motion, the system achieves an ATE of 0.0990\,m and yaw RMSE of 0.0524\,rad, reflecting well-conditioned local estimation. Introducing two $180^\circ$ rotations increases ATE to 0.2917\,m and yaw RMSE to 0.1232\,rad, consistent with weak lateral and heading observability in planar, down-looking configurations. Adding a starfish landmark during rotation improves yaw stability (0.0867\,rad) but further increases ATE (0.3514\,m), indicating that a single visual landmark does not resolve global planar degeneracy. In contrast, introducing depth variation reduces ATE to 0.1954\,m and yaw RMSE to 0.0639\,rad, demonstrating improved conditioning through vertical excitation. The combination of depth variation and landmark constraints yields the best performance (ATE 0.1651\,m, yaw RMSE 0.0412\,rad), confirming that complementary geometric constraints enhance both global and heading consistency. Across all cases, relative pose errors remain low, indicating stable short-horizon estimation despite variations in global drift.

\subsubsection{First Tank: Ablation Study}

An ablation study on the no-rotation trajectory evaluated the contribution of sensing modalities and local refinement. Starting from the full configuration (IMU, DVL, magnetometer, pressure, camera, sliding-window BA), sensors were selectively removed and key parameters varied, including IMU preintegration aggregation and BA window length.
Performance was assessed using normalized ATE, RPE, Yaw$_\text{RMSE}$, and Yaw$_\text{RPE}$. Configurations were compared via Pareto-based multi-objective analysis~\cite{pareto}, where a solution is Pareto-optimal if no other configuration improves one metric without degrading another:
\[
m_j^k \le m_i^k \ \forall k, \quad \text{and} \quad m_j^k < m_i^k \ \text{for at least one } k.
\]
Ranking by Euclidean distance to the ideal point shows that the full configuration (\texttt{FG-UCAL - Full}, 0.6909) outperforms \texttt{FG-UCAL (LBA, 25-Pose Window} (0.7603) and \texttt{FG-UCAL No Local Bundle Adjustment} (0.8075). Removing visual refinement or weakening sensor coupling consistently degrades performance, confirming the benefit of tightly coupled multi-sensor fusion with local optimization.

\begin{figure*}[t]
    \centering
    \includegraphics[width=\linewidth]{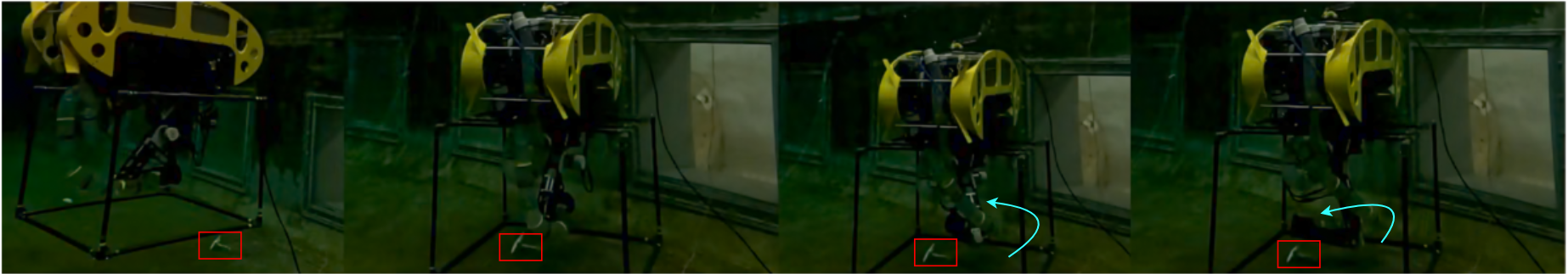}
    \caption{Sequence showing the robot detecting the starfish, approaching it and use the suction-cup to interact with it.}
    \label{fig:snapshot}
\end{figure*}

\begin{figure*}[t]
    \centering
    \includegraphics[width=\linewidth]{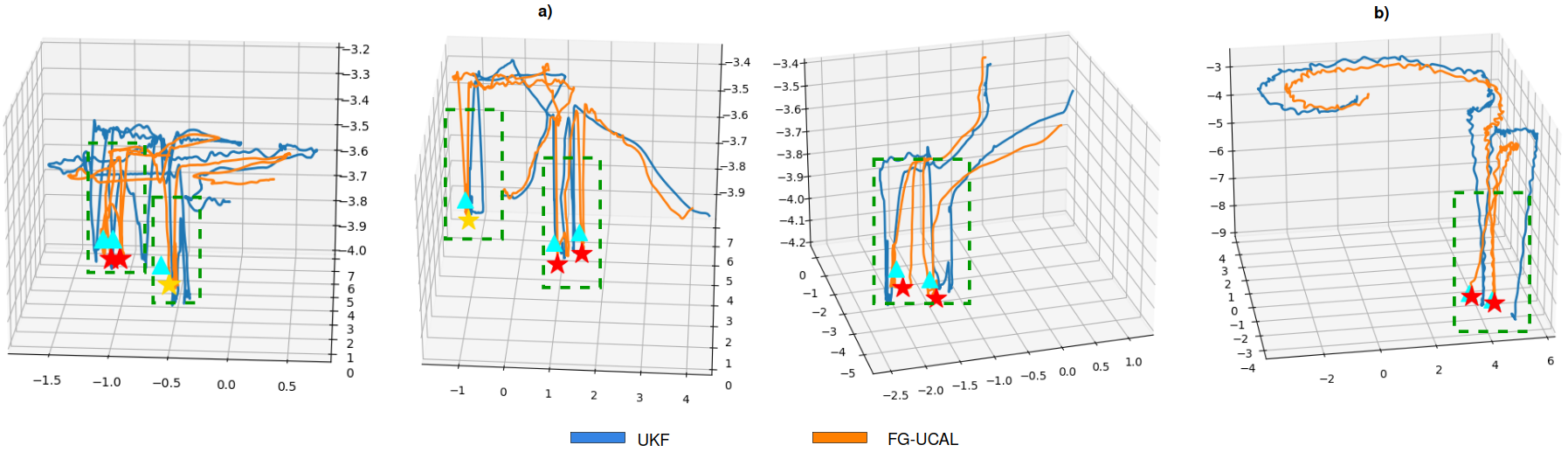}
    \caption{\textbf{a) Second tank experiments}: left) ST1: 2 starfish - 2 pump activation on the last, center)  ST2: 2 starfish - 2 pump activation on the first, right)  ST3: 1 starfish - 2 pump activations. \textbf{b) Harbor experiment}, employing semi-circular trajectory plus 2 dives. Red and gold star markers denote two different starfish targets. Two red stars indicate two separate dives starting from different locations; Cyan triangles indicate suction-pump activation events, and dashed green lines highlight trajectory segments with improved estimation accuracy. }
    \label{fig:jamstec_and_harbor}
\end{figure*}

\begin{table*}[t]
\centering
\footnotesize
\caption{Trajectory accuracy metrics for the Second Tank and Harbor experiments using the full system configuration. Metrics include Absolute Trajectory Error (ATE), Relative Pose Error (RPE), Yaw RMSE, and Yaw RPE.}
\begin{tabular}{lcccc}
\hline
Trajectory & ATE (m) & RPE (m) & Yaw RMSE (rad) & Yaw RPE (rad) \\
\hline

\multicolumn{5}{l}{\textbf{Second Tank Experiments}} \\
\hline

ST1: 2 starfish – 2 pump activation (last)
& 1.0158 & 0.0035 & 0.1836 & 0.0059 \\

ST2: 2 starfish – 2 pump activation (first)
& 0.4985 & 0.0064 & 0.2935 & 0.0303 \\

ST3: 1 starfish – 2 pump activation
& 0.4464 & 0.0152 & 0.1411 & 0.0045 \\

\quad $\hookrightarrow$ No landmark – Starfish revisit distance (m)
& \multicolumn{4}{c}{0.58m} \\

\quad $\hookrightarrow$ Landmark – Starfish revisit distance (m)
& \multicolumn{4}{c}{0.50m} \\

\quad $\hookrightarrow$ Landmark and Pump – Starfish revisit distance (m)
& \multicolumn{4}{c}{0.45m} \\

\hline
\multicolumn{5}{l}{\textbf{Harbor Experiment}} \\
\hline

Harbor
& 0.8223	& 0.0191	& 0.2121	& 0.0751 \\

\quad $\hookrightarrow$ No landmark – Starfish revisit distance (m)
& \multicolumn{4}{c}{1.23m} \\

\quad $\hookrightarrow$ Landmark – Starfish revisit distance (m)
& \multicolumn{4}{c}{1.12m} \\

\quad $\hookrightarrow$ Landmark and Pump – Starfish revisit distance (m)
& \multicolumn{4}{c}{0.90m} \\

\hline
\end{tabular}
\label{tab:trajectory_summary_combined}
\vspace{-1.5em}
\end{table*}

\subsection{Stonefish Simulation: Ground-Truth Validation}

Absolute accuracy was evaluated in Stonefish simulation, where perfect odometry provides ground-truth trajectories. Figure \ref{fig:stonefish_trajs} and Table~\ref{tab:trajectory_summary} report the trajectories along with the ATE, RPE, and yaw errors for representative motion patterns. For well-conditioned motions (forward–lateral and depth–yaw excitation), ATE remains below 0.1\,m. Multiple rotations increase ATE moderately (0.1635\,m), while relative pose errors remain small, indicating strong local consistency. Continuous planar rotations produce the largest drift (ATE 0.998\,m), consistent with reduced observability under repeated yaw motion; however, RPE remains low (0.0032\,m), confirming stable short-horizon estimation.
In the semi-8 trajectory with two starfish dives, global accuracy remains comparable to depth–yaw excitation, and landmark plus pump constraints yield a revisit error of 0.06\,m relative to ground truth, demonstrating metrically consistent anchoring at interaction sites. Overall, drift behavior aligns with the expected geometric conditioning of each motion pattern.

\subsection{Second Tank: Landmark and Contact Evaluation}

Figure~\ref{fig:snapshot} shows a representative starfish interaction sequence in which the vehicle detects the target, approaches it, and activates the suction pump, thereby introducing additional geometric constraints into the factor graph. Figure~\ref{fig:jamstec_and_harbor}a shows the executed trajectories compared against the INS reference. Quantitative results for repeated interaction patterns are summarized in Table~\ref{tab:trajectory_summary_combined}. Global ATE ranges from 0.4464\,m to 1.0158\,m depending on trajectory length and rotational excitation, while relative pose errors remain low (0.0035\,m–0.0152\,m), indicating stable short-horizon estimation despite global drift. Task-level revisit distance provides a more direct measure of mission-relevant spatial consistency. Without landmarks, revisit error is 0.58\,m; adding visual landmark constraints reduces it to 0.50\,m; incorporating both landmarks and pump activation further reduces it to 0.45\,m. The monotonic decrease confirms that contact-induced constraints tighten spatial consistency beyond visual landmarks alone, correcting accumulated drift at physically interacted locations.

\vspace{-0.3em}
\subsection{Harbor Deployment}

The harbor deployment introduces turbidity, illumination variability, and environmental disturbances representative of real-world conditions. The system achieves an ATE of 0.8223\,m and yaw RMSE of 0.2121\,rad. While global drift increases relative to controlled tank trials, task-level revisit accuracy improves as additional constraints are incorporated. Revisit error decreases from 1.23\,m without landmarks to 1.12\,m with visual landmarks, and further to 0.90\,m when pump activation events are included. This reduction demonstrates that contact-induced constraints provide structurally informative corrections even under visually degraded conditions. Improvements are consistent across different dive sequences and initializations. Figure~\ref{fig:jamstec_and_harbor}b compares the executed trajectories against the INS reference.

\section{Discussion and Conclusion}
We presented a contact-aided factor-graph framework in which physical interaction helps resolve degeneracy in perception-limited underwater SLAM. Low-altitude, down-looking operation over planar and texture-sparse seabeds is intrinsically ill-conditioned: lateral motion and yaw are weakly observable, scale depends on depth excitation, and visual landmarks are intermittent. Our formulation tightly couples inertial, DVL, pressure, magnetometer, and visual sensing with physically grounded contact constraints. Experiments in simulation, two tanks, and a harbor deployment show that planar rotations amplify drift, while vertical excitation improves scale and heading observability. Visual landmarks mainly stabilize orientation, but do not fully remove drift over near-planar surfaces. In contrast, contact constraints consistently reduce revisit error and improve spatial consistency at manipulation sites. This trend also holds in harbor conditions, where turbidity and illumination variability increase global ATE, but relative pose errors remain low and contact events provide repeatable corrections. These results show that drift in perception-limited underwater missions is not only a sensing problem, but an observability problem. Structured motion and deliberate physical interaction introduce metrically anchored constraints that improve consistency where manipulation occurs. Future work will extend this framework toward dense, mission-time mapping by coupling contact-aided localization with high-rate volumetric approaches such as FRAGG-MAP~\cite{10801590}, enabling drift-aware reconstruction of benthic workspaces for occupancy, frontier, and manipulation planning.

\bibliographystyle{IEEEtran}
\bibliography{biblio}

\end{document}